\documentclass[10pt]{article}

\usepackage[margin=1.7in]{geometry}
\usepackage{graphicx}
\usepackage[hidelinks]{hyperref}
\graphicspath{{img/}}

\usepackage[frozencache]{minted}

\usepackage[document]{ragged2e}
\usepackage{parskip}

\usepackage[natbibapa]{apacite}

\usepackage[charter]{mathdesign}
\usepackage[scale=0.85]{FiraMono}
\usepackage[semibold]{FiraSans}
\usepackage[T1]{fontenc}

\usepackage{microtype}

\usepackage{caption}
\DeclareCaptionFont{cap}{\firalight}
\author{C. Daniel Greenidge, Noam Miller, and Kenneth A. Norman}
\title{Leabra7: a Python package for modeling recurrent, biologically-realistic neural networks}

\begin{document}
\maketitle

\begin{abstract}
  Emergent \citep{aisaEmergentNeuralModeling2008} is a software
  package that uses the AdEx neural dynamics model
  \citep{bretteAdaptiveExponentialIntegrateandFire2005} and LEABRA
  learning algorithm \citep{oreillyLEABRAModelNeural1996} to
  simulate and train arbitrary recurrent neural network architectures
  in a biologically-realistic manner. We present Leabra7
  \citep{greenidgeCdgreenidgeLeabra7V02018}, a complementary Python
  library that implements these same algorithms. Leabra7 is developed
  and distributed using modern software development principles, and
  integrates tightly with Python's scientific stack. We demonstrate
  recurrent Leabra7 networks using traditional pattern-association
  tasks and a standard machine learning task, classifying the IRIS
  dataset.
\end{abstract}

\tableofcontents{}

\section{Introduction}
Advances in both cognitive modeling and machine learning research will
depend on the ability to simulate and train recurrent neural
networks. Nearly all brain areas contain recurrent
circuitry \citep{shuTurningRecurrentBalanced2003}, from the visual
system \citep{fellemanDistributedHierarchicalProcessing1991} to
the hypothesized canonical cortical microcircuit that grounds
higher-order brain function
\citep{douglasMappingMatrixWays2007}. Biologically plausible
cognitive models must be able to simulate and train similarly recurrent
architectures.

Deep learning is likewise dependent on recurrence. Recurrent neural
networks commonly break records in sequence learning problems such as
language translation
\citep{schmidhuberDeepLearningNeural2015}. However, virtually all
of these networks must use a specialized LSTM architecture to
circumvent backpropagation's vanishing gradient problem. A neural
network model and learning algorithm that could handle arbitrary
feedback connections would be able to take advantage of increasingly
brain-like architectures, improving performance and computational
efficiency \citep{schmidhuberDeepLearningNeural2015}.

Emergent \citep{aisaEmergentNeuralModeling2008} is a cognitive
modeling framework designed to simulate and train these recurrent
architectures. It does this by combining a biologically-plausible
``adaptive-exponential'' (AdEx) neural dynamics model
\citep{bretteAdaptiveExponentialIntegrateandFire2005} with LEABRA,
a local, error-driven, biologically-realistic learning algorithm
\citep{oreillyLEABRAModelNeural1996}. It has been used extensively
by researchers to model the hippocampus
\citep{schapiroComplementaryLearningSystems2017}, prefrontal cortex
and basal ganglia \citep{hazyExecutiveHomunculusComputational2007},
and visual cortex \citep{oreillyRecurrentProcessingObject2013},
among other areas. In each case, the AdEx and LEABRA algorithms
replicate important features of observed neural activity.

Leabra7 \citep{greenidgeCdgreenidgeLeabra7V02018} is a new Python
package that implements Emergent's AdEx neural dynamics model and
LEABRA learning algorithm. It allows researchers to quickly specify
complex cognitive models using the full expressiveness of the Python
programming language. Additionally, since Leabra7 is fully open-source
and contains no compiled code, researchers can easily explore
modifications to the LEABRA algorithm without a detailed knowledge of
programming systems.

Leabra7 is developed and distributed with modern software engineering
practices. It uses an event-driven architecture for low coupling,
extensibility, and, in the future, parallel processing. Continuous
integration, static analysis, and test-driven development ensure code
quality. Both Leabra7 and its dependencies are continuously built and
deployed to Anaconda Cloud, so users can install development and
stable versions of Leabra7 via the \texttt{conda} package manager on
Linux, MacOS, and Windows.

Finally, Leabra7 is deeply integrated with the scientific Python
ecosystem. It natively operates on \texttt{numpy} arrays and
\texttt{pandas} dataframes, and outputs data in a tidy format
\citep{wickhamhadleyTidyData2014}, allowing downstream analysis
using statistical best practices. Since it is distributed with
\texttt{conda}, it is easy to leverage technologies for reproducible
research, such as containers and Jupyter notebooks.

\section{A brief tour of Leabra7's API}

\subsection{Networks, layers, and projections}
The \texttt{Net} (network) object is the primary component of any
Leabra7 simulation: it manages every other object in the network and
the interactions between them. Each simulation begins with the
creation of the \texttt{Net} object and the addition of a few layers:
\begin{minted}{python}
    import leabra7 as lb

    net = lb.Net()
    net.new_layer(name="input", size=3)
    net.new_layer(name="output", size=3)
\end{minted}
Here, the network has two layers, each containing three units, which
are roughly equivalent to neurons.

Connections between layers are managed using \texttt{Projn}
(projection) objects, which are simply bundles of unit-to-unit
connections. Projections can have different connectivity patterns, but
the most common is the ``full'' projection, where every unit in the
sending layer connects to every unit in the receiving layer. Adding a projection
between the input and output layers is accomplished with:
\begin{minted}{python}
    net.new_projn(name="input_to_output",
                  pre="input",
                  post="output")
\end{minted}
Layers are referred to by their names, so that users do not have to
touch the internal representation.

Once the network is constructed, time can be advanced using any of the following three methods:
\begin{minted}{python}
    net.cycle()
    net.minus_phase_cycle(num_cycles=50)
    net.plus_phase_cycle(num_cycles=25)
\end{minted}
The first method simply steps the network forward in time. The last
two methods run a series of cycles after triggering the \emph{minus}
or \emph{plus} phases necessary for the LEABRA learning algorithm
(see~\cite{oreillyComputationalCognitiveNeuroscience2012}, for more
information).

Finally, to compute and apply the weight changes using the LEABRA, the
following method must be called:
\begin{minted}{python}
    net.learn()
\end{minted}

\subsection{Spec objects}
The AdEx neural dynamics model and LEABRA learning algorithm contain
many parameters. Like Emergent, Leabra7 manages parameters through
\texttt{Spec} (specification) objects, which are simple record
classes. Reasonable default parameters are provided automatically, but
can be overridden by providing a custom \texttt{Spec} object during
network creation. For example, to change a layer's inhibition from the
default feedforward/feedback to k-winner-take-all, the following code
can be used:
\begin{minted}{python}
    import leabra7 as lb

    net = lb.Net()
    net.new_layer(name="input",
                  size=3,
                  spec=lb.LayerSpec(inhibition_type="kwta"))
\end{minted}
As much as possible, Leabra7 retains the names given to parameters in
the Emergent software.

\subsection{Output}

Leabra7 outputs data in two forms: observations, which are
instantaneous snapshots of the network state, and logs, which are
series of observations recorded over time.

Observations are requested using the \texttt{Net.observe()} method. For
example, the following code observes the output layer activation:
\begin{minted}{python}
    >> net.observe(name="output", attr="unit_act")
         unit  act
    0     0  0.0
    1     1  0.0
    2     2  0.0
\end{minted}
Here, the \texttt{name} parameter is the name of the object to
observe, the \texttt{attr} parameter is the object attribute to
observe, and the output is a Pandas dataframe. Observations can be
requested for any attribute, on any object, at any time.

On the other hand, logging over time must be requested at network
instantiation, through a \texttt{Spec} opbject. Here is an example of
logging the \texttt{output} layer's unit activation every cycle:
\begin{minted}{python}
    layer_spec = lb.LayerSpec(
        log_on_cycle=("unit_act", )
    )
    net.new_layer(name="output", size=3, spec=layer_spec)
\end{minted}
After a few cycles, the logs can be requested:
\begin{minted}{python}
    >> logs = net.logs(freq="cycle", name="output")
\end{minted}
Because the activation is an attribute of the \emph{parts} of
the output layer, not of the entire layer itself (like the average activation), we
access the \texttt{parts} member of the returned tuple:
\begin{minted}{python}
    >> logs.parts
        unit           act  time
    6      0  1.022608e-20     1
    7      1  1.022608e-20     1
    8      2  1.022608e-20     1
    9      0  1.181225e-20     2
    10     1  1.181225e-20     2
    11     2  1.181225e-20     2
\end{minted}
As before, the logs are stored in a Pandas dataframe, except that here
the \texttt{time} column denotes the cycle at which the activation was
recorded.

Since logging slows down network cycling and takes a
considerable amount of memory, it is typically disabled during
training and re-enabled only when needed, using the
\texttt{Net.pause\_logging()} and \texttt{Net.resume\_logging()}
methods.

\section{Example networks}

\subsection{Two neurons}\label{sec:two-neurons}
\begin{figure}[t]
    \centering
    \includegraphics[width=5in, height=3in]{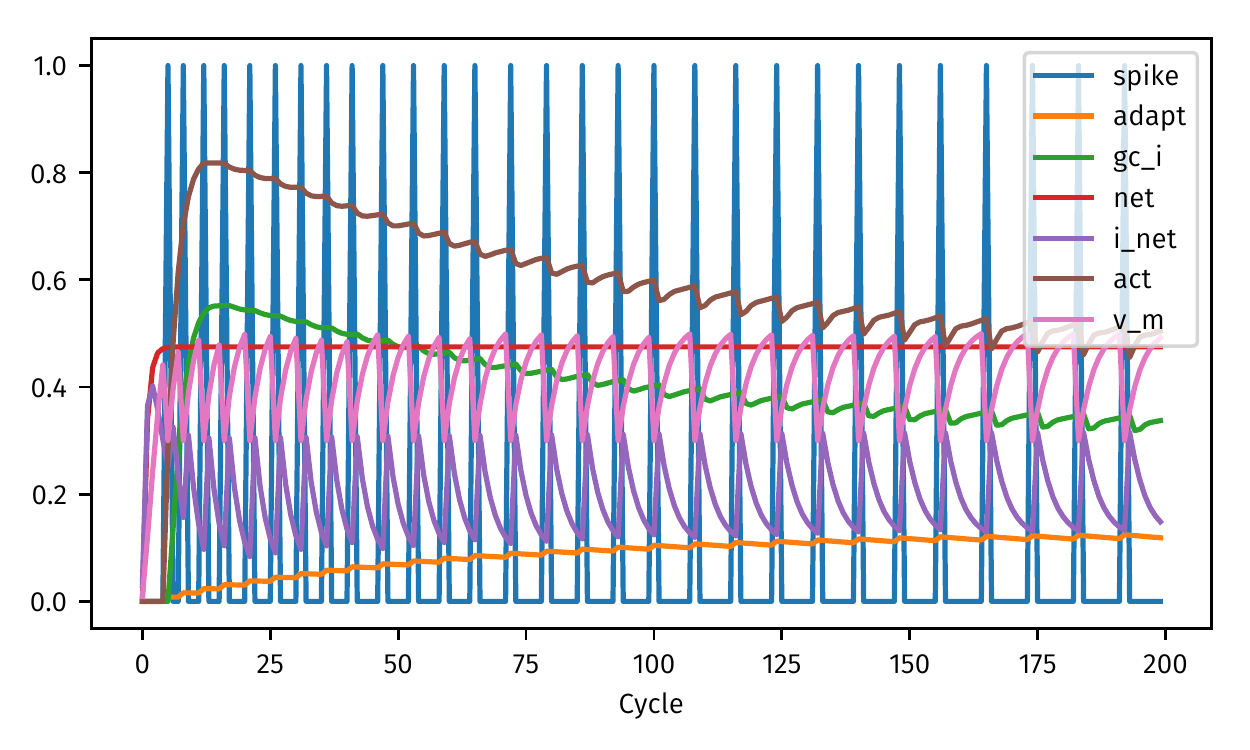}
    \caption{The internal dynamics of a spiking unit.
      See \S\ref{sec:two-neurons} for a detailed
      explanation of each quantity.}\label{fig:two-neurons}
\end{figure}

In the most basic network architecture, a single input unit is
connected to a single output unit, with a connection weight of
\(0.5\). The input unit's activation is clamped at a high value of
\(0.95\), so that it excites repeated spiking in the the output unit.

The dynamics of the output unit are shown in
Figure~\ref{fig:two-neurons}. As the time-integrated excitatory input
\texttt{net} rises, so does the net current into the unit,
\texttt{i\_net}. This drives the unit's electric potential,
\texttt{v\_m}, above the spiking threshold, causing repeated
spiking. When the unit spikes, the rate-coded activation value
\texttt{act} increases, which would then be transmited to downstream,
postsynaptic neurons if any existed. The spiking triggers feedback
inhibition, \texttt{gc\_i}, from the layer. Over time, the unit's
adaption current \texttt{adapt} increases, representing the onset of a
refractory period, and spiking slows. When spiking slows, the
activation \texttt{act} and inhibition \texttt{gc\_i} drop.

\subsection{Pattern association}

\begin{figure}[t]
    \centering
    \includegraphics[width=5in, height=3.25in]{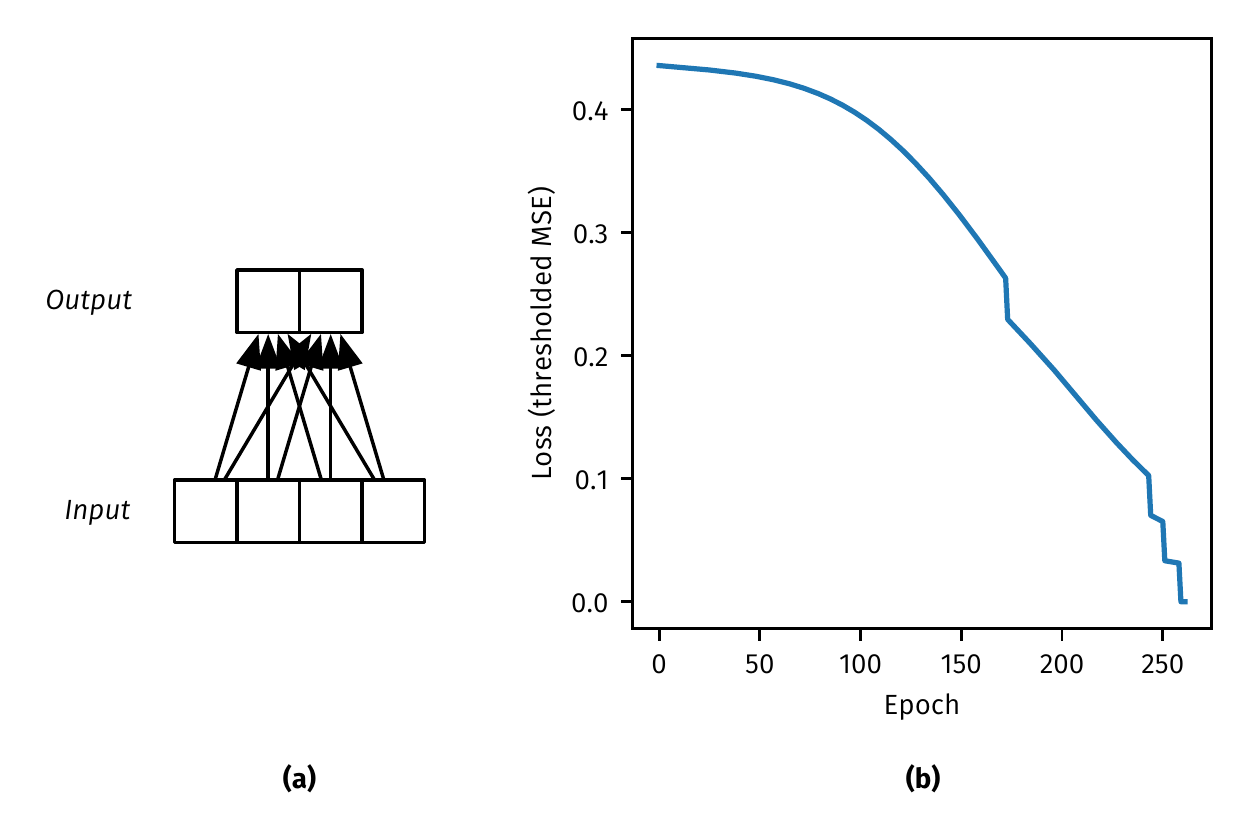}
    \caption{\textbf{(a)} The architecture of the pattern association
      network. Arrows represent connections between units. A size-2
      input layer is connected to a size-4 output layer with a
      fully-connected projection. There is no hidden
      layer. \textbf{(b)} The thresholded mean-squared-error (MSE)
      loss during training, over 500 epochs. Thresholded MSE is
      similar to standard MSE, but errors less 0.5 are set to 0 (so
      the output activation only has to be on the correct side of
      0.5). Because this is a strictly feedforward network, the
      loss curve decreases roughly
      monotonically.}\label{fig:pat-assoc}
\end{figure}

A simple input-output architecture with no hidden layer (see
Figure~\ref{fig:pat-assoc}a) can be trained to recognize the following
patterns, in which active units are dark:
\begin{center}
    \includegraphics[width=0.3\textwidth]{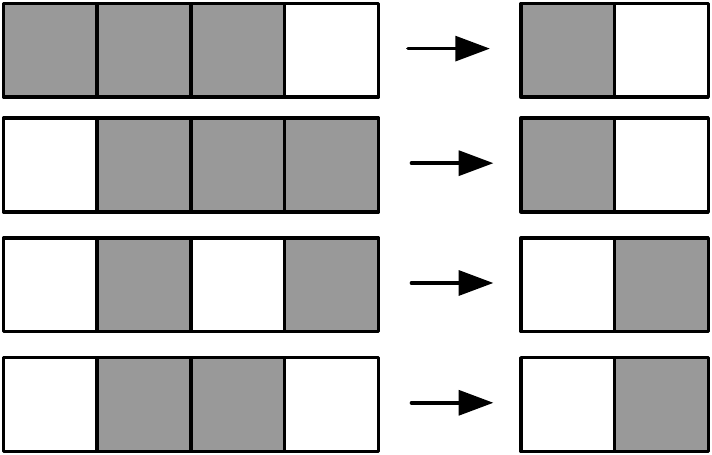}
\end{center}
To train the network, the input pattern is clamped to the network's
input layer, i.e.\ the input layer's activations are set manually to
the input pattern. The network is then cycled 50 to 75 times, to
generate an output response. This is known as the ``minus''
phase. Then, in the ``plus'' phase, the correct output pattern is
clamped to the output layer for an additional 20 to 25
cycles. Following the plus phase, the LEABRA algorithm is used to
compute and apply weight changes that will reduce the error between
the minus and plus phases. This procedure is repeated for each pattern
in the training set, forming an epoch. The network loss over 500
training epochs is shown in Figure~\ref{fig:pat-assoc}b.

\subsection{Error-driven hidden}
\begin{figure}[t]
    \centering
    \includegraphics[width=5in, height=3.25in]{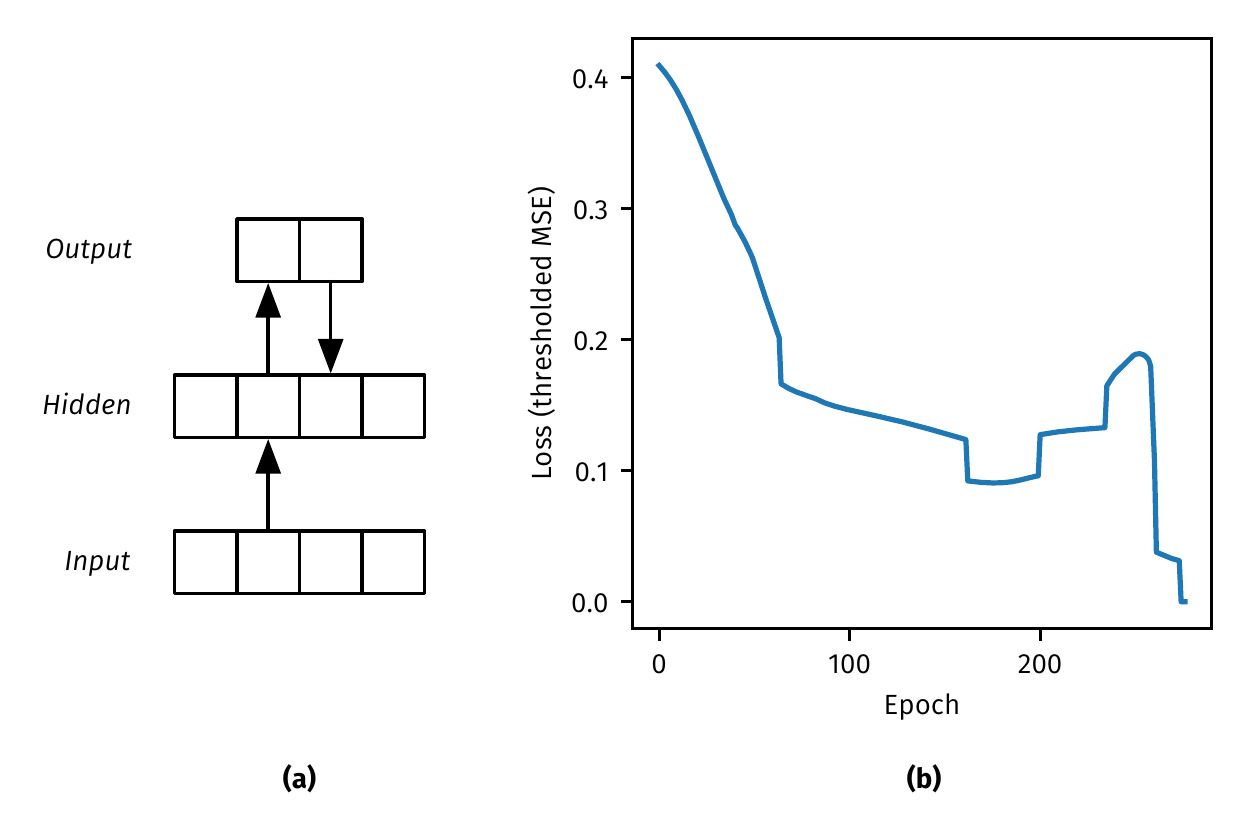}
    \caption{\textbf{(a)} The architecture of the error-driven hidden
      network. Arrows represent ``projections'', which are groups of
      connections (in this case, connections which go from every unit
      in the sending layer to every unit in the receiving layer.) A
      recurrent feedback projection runs between the output and hidden
      layers to provide an error signal. \textbf{(b)} The thresholded
      sum-of-squared-error loss during training. The feedback
      connections induce high sensitivity to even tiny weight
      changes, causing oscillations in the loss curve. Such
      oscillations are not observed with feedforward
      architectures---see
      Figure~\ref{fig:pat-assoc}.}\label{fig:err-hidden}
\end{figure}

The following patterns are an example of a \emph{nonlinear
  discrimination problem} that cannot be learned using the simple
pattern association architecture
\citep{oreillyComputationalPrinciplesLearning2000}:
\begin{center}
    \includegraphics[width=0.3\textwidth]{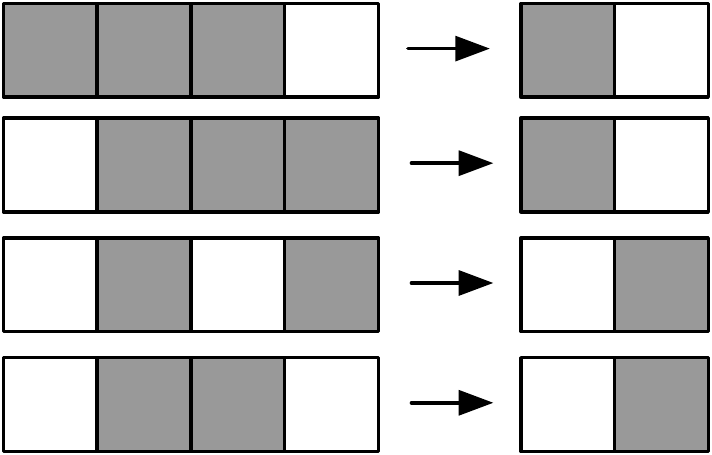}
\end{center}

Intuitively, the simple architecture cannot learn the patterns because
each input unit is active an equal amount for each output
pattern. Mathematically, it is because the data are not linearly
separable in \(\mathbb{R}^4\).  Introducing a hidden layer allows the
network to solve such problems (Figure~\ref{fig:err-hidden}). A
feedback connection is also required to provide an error response to
the hidden layer during the plus phase of learning.

\subsection{Classifying the IRIS dataset}
\begin{figure}[t]
    \centering
    \includegraphics[width=4.5in, height=3in]{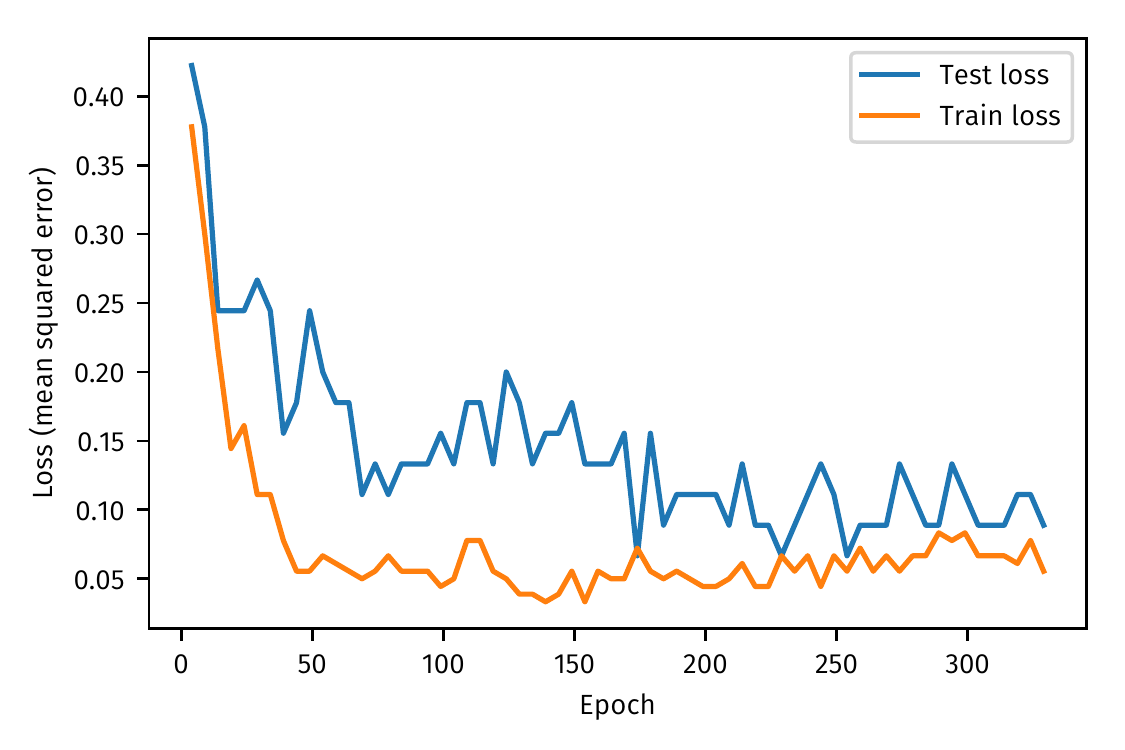}
    \caption{Training and test loss (mean squared error) for an
      input-hidden-output IRIS classifier. Network output was
      binarized by setting the largest value to 1 and all other values
      to 0. After 330 epochs, training classification accuracy was
      95.83\% and test classification accuracy was 90.00\%.
    }\label{fig:iris-loss}
\end{figure}
The IRIS dataset is a standard machine learning classification
benchmark \citep{fisherUseMultipleMeasurements1936} containing 150
examples. The data consist of four input features: sepal length, sepal
width, petal length, and petal width. The output class is one of three
iris flower species: setosa, versicolor, or virginica.

We encoded the continuous input variables as follows: first, for each
input value \(x \in \mathbb{R}\), we estimated its quantile \(q(x)\),
which maps \(x\) into the interval \([0, 1]\). Next, we broke the
\([0, 1]\) interval into \(n\) bins and recorded the index \(i\) of
the bin into which \(q(x)\) falls. Finally, we encoded \(i\) using a
one-hot encoding, producing a vector with each value in \(\{0, 1\}\).

After preprocessing the data, we classified IRIS using the same
input-hidden-output architecture described in
Fig~\ref{fig:err-hidden}. Using \(n=10\) bins, an input layer size of
36, and a hidden layer size of 23 produced satisfactory results. After
500 epochs of training on 80\% of the dataset, we obtained a training
accuracy of 95.83\%, with a test accuracy of 90.00\% on the remaining
20\% (Figure~\ref{fig:iris-loss}).

\section{Conclusion and future work}
Leabra7 is a Pythonic, modern implementation of the LEABRA algorithm
first implemented in Emergent
\citep{aisaEmergentNeuralModeling2008}. It is easy to modify, and
integrates seamlessly into modern scientific workflows. We have shown
that Leabra7 networks can learn both traditional pattern association
tasks and standard machine learning tasks using feedforward and
recurrent architectures as appropriate.

Before Leabra7 can be used to implement large cognitive models, like
the hippocampal model described in
\cite{schapiroComplementaryLearningSystems2017}, performance must
be improved in at least two areas. The first is learning performance:
adaptive learning rate adjustment and annealing would greatly minimize
late-stage loss oscillations. Second, a compiled engine in a language
such as C++ would improve training speed for large networks, at the
cost of making the algorithm more difficult to modify.

\bibliographystyle{apacite}
\bibliography{library.bib}

\appendix{}

\section{Algorithm pseudocode}
This section documents the exact algorithm that Leabra7 implements, as
of v0.1.0. For a detailed explanation of the algorithm, see
\cite{oreillyComputationalCognitiveNeuroscience2012}, and for default
parameter values, see the file \texttt{leabra7/specs.py} in Leabra7's
source code. Nonessential features such as relative input scaling and
Hebbian threshold modulation are omitted.

\subsection{Network cycling}
In a network cycle, the differential equations that govern the network
dynamics are integrated forwards one step in time. This process
consists of two stages. First, in the activation cycle, each unit is
advanced one step forward in time. Typically, this causes the units to
change their activation value. Then, in the projection flush, sending
units propagate their new activations to receiving units.

\begin{enumerate}

\item Activation cycle

  For each layer, perform the following steps:

  \begin{enumerate}

  \item For each unit, update net input.

\begin{verbatim}
unit.net += integ * net_dt * (sum(inputs) - unit.net)
\end{verbatim}

    \texttt{unit.net} is the unit's net input, \texttt{inputs}
    is an array of activations sent to this unit in the projection
    flushing stage, \texttt{integ} is the global integration time
    constant, and \texttt{net\_dt} is the net input integration time
    constant.

  \item Update layer-wide inhibition. These values are calculated at
    the layer level, but they are needed at the unit level for
    updating the membrane potential.

    Feedforward inhibition (\texttt{ffi}):
\begin{verbatim}
ffi = ff * max(layer.avg_net - ff0, 0)
\end{verbatim}
    \texttt{ff} is the feedforward inhibition multiplier, which
    controls the strength of feedforward inhibition, \texttt{ff0} is
    the feedforward inhibition offset, and \texttt{layer.avg\_net} is
    the average net input across the layer.

    Feedback inhibition (\texttt{fbi}):
\begin{verbatim}
layer.fbi += fb_dt * (fb * net.avg_act - layer.fbi)
\end{verbatim}
    \texttt{fb} is the feedback inhibition multiplier, which controls
    the strength of feedback inhibition, and \texttt{fb\_dt} is the
    feedback inhibition integration time constant.

    Global inhibition (\texttt{gc\_i}):
\begin{verbatim}
layer.gc_i = gi * (ffi * layer.fbi)
\end{verbatim}
    \texttt{gi} is the global inhibition multiplier, which controls
    overall inhibition strength.

  \item For each unit, update the net current \texttt{i\_net} and
    membrane potential \texttt{v\_m}.

    We maintain two versions of the net current and membrane
    potential: rate-coded, which approximates the discrete spiking
    rate with a number in the interval \([0, 1]\), and
    non-rate-coded. The rate-coded version does not reset after each
    spike, and the non-rate-coded version does reset.

    First, calculate the non-rate-coded net current and membrane potential:
\begin{verbatim}
unit.i_net = (unit.net * (e_rev_e - unit.v_m) +
              gc_l * (e_rev_l - unit.v_m) +
              layer.gc_i * (e_rev_i - unit.v_m))
unit.v_m += clamp(integ * vm_dt * unit.i_net,
                  min=-100, max=100)
\end{verbatim}
    We limit changes in \texttt{v\_m} to be in the range \([-100, 100]\) to
avoid numerical integration problems.  Here, \texttt{e\_rev\_e} is the
excitatory reversal potential, \texttt{e\_rev\_l} is the leak reversal
potential, and \texttt{e\_rev\_i} is the inhibitory reversal
potential. \texttt{gc\_l} is the leak conductance, which always remains
constant.

    Next, calculate the rate-coded membrane potential, which does not reset when the unit spikes:
\begin{verbatim}
unit.v_m_eq += clamp(integ * vm_dt * unit.i_net,
                     min=-100, max=100)
\end{verbatim}
    The \texttt{eq} suffix denotes ``equilibrium''.

  \item Update activation.

    Calculate the excitatory conductance (net input) that would place the unit
    at the spike threshold \texttt{spk\_thr}:
\begin{verbatim}
g_e_thr = (layer.gc_i * (e_rev_i - spk_thr) +
           gc_l * (e_rev_l - spk_thr) -
           unit.adapt) / (spk_thr - e_rev_e)
\end{verbatim}
    \texttt{adapt} is the unit's adaption current, which increases
    slowly as the unit activates and depresses the spike rate.

    Next, we handle discrete spiking. Discrete spiking is always
    calculated even with rate-coded activation, since adaption current
    depends on it.
\begin{verbatim}
if unit.v_m > spk_thr:
    unit.v_m = v_m_r
    unit.spike = 1
else:
    unit.spike = 0
\end{verbatim}
    \texttt{v\_m\_r} is the reset value for the membrane potential after a spike.

    The unit activation can now be calculated and integrated forward in time:
\begin{verbatim}
if unit.v_m_eq < unit.spk_thr:
    new_act = nxx1(unit.v_m_eq - spk_thr)
else:
    new_act = nxx1(unit.net - g_e_thr)
unit.act += integ * vm_dt * (new_act - unit.act)
\end{verbatim}
    Before the first ``spike'', the activation is governed by the
    \texttt{v\_m\_eq} dynamics, but after the first spike, it is
    driven by the \texttt{g\_e\_thr} term. \texttt{nxx1} is the noisy
    \(X / (X + 1)\) activation function described
    in~\cite{oreillyComputationalCognitiveNeuroscience2012}.

    Finally, the adaption current is calculated:
\begin{verbatim}
unit.adapt += integ * adapt_dt * vm_gain * (vm_gain * (
    unit.v_m - e_rev_l)) + lunit.spike * spike_gain
\end{verbatim}
    As usual, \texttt{adapt\_dt} is the adaption current integration
    time constant. The constant \texttt{vm\_gain} controls how much
    the membrane potential increases adaption current, and the
    constant \texttt{spike\_gain} controls how much discrete spikes
    increase the adaption current.

  \item Update cycle learning averages.

    The LEABRA algorithm drives learning off of cascading activation
    averages: In general, the short-term averages encode the expected,
    or ``plus-phase'' value, while the medium-term average encodes the
    network's current response, or ``minus-phase'' value mixed with
    the plus-phase value.

    The supershort, short, and medium-term learning averages that
    drive error-based learning are calculated every cycle:
\begin{verbatim}
unit.avg_ss += integ * ss_dt * (
    unit.act - unit.avg_ss)
unit.avg_s += integ * s_dt * (
    unit.avg_ss - unit.avg_s)
unit.avg_m += integ * m_dt * (
    unit.avg_s - unit.avg_m)
\end{verbatim}
    \texttt{ss\_dt}, \texttt{s\_dt}, and \texttt{m\_dt} are the
    respective integration time constants.

    Every trial, the long-term learning average, which drives Hebbian
    learning, is calculated:
\begin{verbatim}
if unit.avg_m > 0.1:
    unit.avg_l += unit.avg_m * l_up_inc
else:
    unit.avg_l += acts_p_avg * l_dn_dt * (
        unit.avg_m - unit.avg_l)
\end{verbatim}
    Note that the derivative of \texttt{avg\_l} is bounded above
    \(0.1\). \texttt{l\_dn\_dt} is the ``downwards'' integration time
    constant, and \texttt{acts\_p\_avg} is the average plus-phase
    activation across the layer.
  \end{enumerate}

\item Projection flushing.

  Once each unit's activation has been updated, for each connection in
  each projection, we propagate the sending unit's activation to the
  receiving unit, scaled by the connection weight:
\begin{verbatim}
for connection in projection.connections:
    connection.receiving_unit.add_input(
        connection.weight * connection.sending_unit.act
    )
\end{verbatim}
  The \texttt{add\_input} function simply keeps the running sum of every
  added input, which is later integrated in the ``update net input'' step.

  \end{enumerate}

  \subsection{Learning}
  Learning trials consist of a ``minus phase'' of 50--75 cycles,
  followed by ``plus phase'' of 20--25 cycles. In the minus phase, the
  activations of the network's input layers are set to the input
  pattern (also known as ``clamping''), but the output layers are left
  untouched. In the plus phase, the input pattern is still clamped,
  and the desired output pattern is clamped on the network's output
  layers.

  In the following pseudocode, \texttt{pre} refers to the connection's
  sending unit and \texttt{post} refers to the connection's receiving
  unit.

  After each trial, we compute the learning equations. First, the
  short- and medium-term learning coproducts, which encode the plus
  and minus-phase activations, respectively:
\begin{verbatim}
srs = post.avg_s * pre.avg_s
srm = post.avg_m * pre.avg_m
s_mix = 0.9
sm_mix = s_mix * srs + (1 - s_mix) * srm
\end{verbatim}
  Some of the medium-term coproduct is mixed into the short-term
  coproduct to drive weight depression if the receiving unit has \(0\)
  activation in the plus phase.

  Next, we calculate the long-term and medium-term floating
  thresholds, which encode Hebbian and error-driven learning, respectively:
\begin{verbatim}
lthr = post.avg_l * pre.avg_m * thr_l_mix
mthr = srm * (1 - thr_l_mix)
\end{verbatim}
  The constant \texttt{thr\_l\_mix} determines how much learning is
  Hebbian (based on \texttt{lthr}) and how much learning is
  error-driven (based on \texttt{mthr}).

  Now we calculate the change (\texttt{dwt}) in the connection weight
  \texttt{fwt}:
\begin{verbatim}
dwt = lrate * xcal(sm_mix, lthr + mthr)
if dwt > 0:
    dwt *= 1 - fwt
else:
    dwt *= self.fwt
fwt += dwt
\end{verbatim}
  The multiplications ensure that the weight change slows down
  exponentially near \(1\) and \(0\).

  Finally, we calculate \texttt{wt}, the sigmoidal contrast-enhanced
  weight used to propagate activation to the postsynaptic unit:
\begin{verbatim}
wt = 1 / (1 + (sig_offset * (1 - x) / x)^sig_gain)
\end{verbatim}
  The constant \texttt{sig\_offset} controls the sigmoid translation and
  the constant \texttt{sig\_gain} controls the sigmoid gain.

  \setminted{fontsize=\small}
  \newgeometry{centering=True}
  \section{Example network source code}
  \subsection{Two neurons}

\begin{minted}{python}
import matplotlib.pyplot as plt
import pandas as pd  #type: ignore
import leabra7 as lb

# Create the network
net = lb.Net()

# Log internal dynamics for each layer
layer_spec = lb.LayerSpec(
    log_on_cycle=("unit_v_m", "unit_act", "unit_i_net", "unit_net",
                  "unit_gc_i", "unit_adapt", "unit_spike"))

# Instantiate the network
net.new_layer(name="input", size=1, spec=layer_spec)
net.new_layer(name="output", size=1, spec=layer_spec)
net.new_projn(name="proj1", pre="input", post="output")

# Activate the input layer
net.clamp_layer(name="input", acts=[1])

# Cycle the network
for i in range(200):
    net.cycle()

# Retrieve logs
wholeLog, partLog = net.logs(freq="cycle", name="output")

# Plot the logs
ax = partLog.plot(x="time")
ax.set_xlabel("Cycle")
ax.legend(loc="upper right")
ax.figure.tight_layout()
ax.figure.savefig("two_neurons.png")
\end{minted}

  \subsection{Pattern association}

\begin{minted}{python}
import logging
import sys
from typing import Dict
from typing import List
from typing import Iterable
from typing import Tuple
from typing import Union

import numpy as np  # type: ignore
import pandas as pd  # type: ignore

import leabra7 as lb  # type: ignore

LoggerType = Union[None, logging.Logger, logging.LoggerAdapter]


def load_data() -> Tuple[np.ndarray, np.ndarray]:
    """Loads and preprocesses the data.

    Returns:
      An (X, Y) tuple containing the features and labels, respectively.

    """
    # yapf: disable
    X = np.array([
        [1, 1, 1, 0],
        [0, 1, 1, 1],
        [0, 1, 0, 1],
        [0, 1, 0],
    ])
    Y = np.array([
        [1, 0],
        [1, 0],
        [0, 1],
        [0, 1]
    ])
    # yapf: enable
    return (X, Y)


def build_network(logger: LoggerType = None) -> lb.Net:
    """Builds the classifier network.

    Args:
      logger: The logger to use.

    Returns:
      A Leabra7 network for classification.

    """
    if logger is None:
        logger = logging.getLogger()
    logger.info("Building network")
    net = lb.Net()

    # Layers
    layer_spec = lb.LayerSpec(
        gi=1.5,
        ff=1,
        fb=0.5,
        fb_dt=0.7,
        log_on_epoch=("unit_act", ),
        unit_spec=lb.UnitSpec(
            adapt_dt=0,
            vm_gain=0,
            spike_gain=0,
            ss_dt=1,
            s_dt=0.2,
            m_dt=0.15,
            l_dn_dt=0.4,
            l_up_inc=0.15,
            vm_dt=0.3,
            net_dt=0.7))
    net.new_layer("input", size=4, spec=layer_spec)
    net.new_layer("output", size=2, spec=layer_spec)

    # Projections
    spec = lb.ProjnSpec(
        lrate=0.02,
        dist=lb.Uniform(0.25, 0.75),
        thr_l_mix=0.01,
        cos_diff_lrate=False)
    net.new_projn("input_to_output", pre="input", post="output", spec=spec)

    return net


def trial(network: lb.Net, input_pattern: Iterable[float],
          output_pattern: Iterable[float]) -> None:
    """Runs a trial.

    Args:
      input_pattern: The pattern to clamp to the network's input layer.
      output_pattern: The pattern to clamp to the network's output layer.

    """
    network.clamp_layer("input", input_pattern)
    network.minus_phase_cycle(num_cycles=100)
    network.clamp_layer("output", output_pattern)
    network.plus_phase_cycle(num_cycles=20)
    network.unclamp_layer("input")
    network.unclamp_layer("output")
    network.learn()


def epoch(network: lb.Net, input_patterns: np.ndarray,
          output_patterns: np.ndarray) -> None:
    """Runs an epoch (one pass through the whole dataset).

    Args:
      input_patterns: A numpy array with shape (n_samples, n_features).
      output_patterns: A numpy array with shape (n_samples, n_features).

    """
    for x, y in zip(input_patterns, output_patterns):
        trial(network, x, y)
    network.end_epoch()


def mse_thresh(expected: np.ndarray, actual: np.ndarray) -> float:
    """Calculates the thresholded mean squared error.

    If the error is < 0.5, it is treated as 0 (i.e., we count < 0.5 as 0 and
    > 0.5 as 1).

    Args:
      expected: The expected output pattern.
      actual: The actual output pattern

    Returns:
      The thresholded mean squared error.

    """
    diff = np.abs(expected - actual)
    diff[diff < 0.5] = 0
    return np.mean(diff * diff)


def train(network: lb.Net,
          input_patterns: np.ndarray,
          output_patterns: np.ndarray,
          num_epochs: int = 3000,
          logger: LoggerType = None) -> pd.DataFrame:
    """Trains the network.

    Args:
      input_patterns: A numpy array with shape (n_samples, n_features).
      output_patterns: A numpy array with shape (n_samples, n_features).
      num_patterns: The number of epochs to run. Defaults to 500.
      logger: The logger to use. If None, will use the module's default logger.

    Returns:
      pd.DataFrame:  A dataframe of metrics from the training run.

    """
    if logger is None:
        logger = logging.getLogger()
    logger.info("Begin training")

    data: Dict[str, List[float]] = {
        "epoch": [],
        "train_loss": [],
    }

    perfect_epochs = 0
    for i in range(num_epochs):
        epoch(network, input_patterns, output_patterns)
        pred = predict(network, input_patterns)
        data["epoch"].append(i)
        data["train_loss"].append(mse_thresh(output_patterns, pred))

        logger.info("Epoch %d/%d. Train loss: %.4f", i, num_epochs,
                    data["train_loss"][-1])

        if data["train_loss"][-1] == 0:
            perfect_epochs += 1
        else:
            perfect_epochs = 0

        if perfect_epochs == 3:
            logger.info("Ending training after %d perfect epochs.",
                        perfect_epochs)
            break

    return pd.DataFrame(data)


def output(network: lb.Net, pattern: Iterable[float]) -> List[float]:
    """Calculates a prediction for a single input pattern.

    Args:
      network: The trained network.
      pattern: The input pattern.

    Returns:
      np.ndarray: The output of the network after clamping the input
      pattern to the input layer and settling. The max value is set to one,
      everything else is set to zero.

    """
    network.clamp_layer("input", pattern)
    for _ in range(50):
        network.cycle()
    network.unclamp_layer("input")
    out = network.observe("output", "unit_act")["act"].values
    return list(out)


def predict(network: lb.Net, input_patterns: np.ndarray) -> np.ndarray:
    """Calculates predictions for an array of input patterns.

    Args:
        network (lb.Net): The trained network.
        input_patterns (np.ndarray): An array of shape (n_samples, n_features)
            containing the input patterns for which to calculate predictions.

    Returns:
        np.ndarray: An array of shape (n_samples, n_features) containing the
            predictions for the input patterns.

    """
    outputs = []
    for item in input_patterns:
        outputs.append(output(network, item))
    return np.array(outputs)


if __name__ == "__main__":
    PROJ_NAME = "pat_assoc"

    logging.basicConfig(
        level=logging.DEBUG,
        format="%(asctime)s %(levelname)s %(message)s",
        handlers=(logging.FileHandler(
            "{0}_log.txt".format(PROJ_NAME), mode="w"),
                  logging.StreamHandler(sys.stdout)))

    logging.info("Begin training %s", PROJ_NAME)

    X, Y = load_data()
    net = build_network()
    metrics = train(net, X, Y)

    # Save metrics and network for future analysis
    metrics.to_csv("{0}_metrics.csv".format(PROJ_NAME), index=False)
    net.save("{0}_network.pkl".format(PROJ_NAME))
\end{minted}

  \subsection{Error-driven hidden}
\begin{minted}{python}
import logging
import sys
from typing import Dict
from typing import List
from typing import Iterable
from typing import Tuple
from typing import Union

import numpy as np  # type: ignore
import pandas as pd  # type: ignore

import leabra7 as lb  # type: ignore

LoggerType = Union[None, logging.Logger, logging.LoggerAdapter]


def load_data() -> Tuple[np.ndarray, np.ndarray]:
    """Loads and preprocesses the data.

    Returns:
       An (X, Y) tuple containing the features and labels, respectively.

    """
    # yapf: disable
    X = np.array([
        [1, 0, 1, 0],
        [0, 1, 0, 1],
        [1, 1, 0, 0],
        [0, 0, 1, 1,],
    ])
    Y = np.array([
        [1, 0],
        [1, 0],
        [0, 1],
        [0, 1],
    ])
    # yapf: enable
    return (X, Y)


def build_network(logger: LoggerType = None) -> lb.Net:
    """Builds the classifier network.

    Args:
      logger: The logger to use.

    Returns:
      A Leabra7 network for classification.

    """
    if logger is None:
        logger = logging.getLogger()
    logger.info("Building network")
    net = lb.Net()

    # Layers
    layer_spec = lb.LayerSpec(
        gi=1.5,
        fb=1,
        ff=1,
        unit_spec=lb.UnitSpec(
            adapt_dt=0,
            vm_gain=0,
            spike_gain=0,
            ss_dt=1,
            s_dt=0.2,
            m_dt=0.1,
            l_dn_dt=0.4,
            l_up_inc=0.15,
            vm_dt=1 / 3.3,
            net_dt=0.7))
    net.new_layer("input", size=4, spec=layer_spec)
    net.new_layer("hidden", size=4, spec=layer_spec)
    net.new_layer("output", size=2, spec=layer_spec)

    # Projections
    lrate = 0.04  # This is high, but it will be adaptively decreased
    up_spec = lb.ProjnSpec(
        lrate=lrate,
        dist=lb.Uniform(0.25, 0.75),
        thr_l_mix=0,
        cos_diff_lrate=False,
        cos_diff_thr_l_mix=True)
    down_spec = lb.ProjnSpec(
        lrate=lrate,
        dist=lb.Uniform(0.25, 0.75),
        wt_scale_rel=0.3,
        thr_l_mix=0,
        cos_diff_lrate=False,
        cos_diff_thr_l_mix=True)
    net.new_projn("input_to_hidden", pre="input", post="hidden", spec=up_spec)
    net.new_projn(
        "hidden_to_output", pre="hidden", post="output", spec=up_spec)
    net.new_projn(
        "output_to_hidden", pre="output", post="hidden", spec=down_spec)

    return net


def trial(network: lb.Net, input_pattern: Iterable[float],
          output_pattern: Iterable[float]) -> None:
    """Runs a trial.

    Args:
      input_pattern: The pattern to clamp to the network's input layer.
      output_pattern: The pattern to clamp to the network's output layer.

    """
    network.clamp_layer("input", input_pattern)
    network.minus_phase_cycle(num_cycles=50)
    network.clamp_layer("output", output_pattern)
    network.plus_phase_cycle(num_cycles=20)
    network.unclamp_layer("input")
    network.unclamp_layer("output")
    network.learn()


def epoch(network: lb.Net, input_patterns: np.ndarray,
          output_patterns: np.ndarray) -> None:
    """Runs an epoch (one pass through the whole dataset).

    Args:
      input_patterns: A numpy array with shape (n_samples, n_features).
      output_patterns: A numpy array with shape (n_samples, n_features).

    """
    for x, y in zip(input_patterns, output_patterns):
        trial(network, x, y)
    network.end_epoch()


def mse_thresh(expected: np.ndarray, actual: np.ndarray) -> float:
    """Calculates the thresholded mean squared error.

    If the error is < 0.5, it is treated as 0 (i.e., we count < 0.5 as 0 and
    > 0.5 as 1).

    Args:
      expected: The expected output pattern.
      actual: The actual output pattern

    Returns:
      The thresholded mean squared error.

    """
    diff = np.abs(expected - actual)
    diff[diff < 0.5] = 0
    return np.mean(diff * diff)


def train(network: lb.Net,
          input_patterns: np.ndarray,
          output_patterns: np.ndarray,
          num_epochs: int = 3000,
          logger: LoggerType = None) -> pd.DataFrame:
    """Trains the network.

    Args:
      input_patterns: A numpy array with shape (n_samples, n_features).
      output_patterns: A numpy array with shape (n_samples, n_features).
      num_patterns: The number of epochs to run. Defaults to 500.
      logger: The logger to use. If None, will use the module's default logger.

    Returns:
      pd.DataFrame:  A dataframe of metrics from the training run.

    """
    if logger is None:
        logger = logging.getLogger()
    logger.info("Begin training")

    data: Dict[str, List[float]] = {
        "epoch": [],
        "train_loss": [],
    }

    perfect_epochs = 0
    for i in range(num_epochs):
        epoch(network, input_patterns, output_patterns)
        pred = predict(network, input_patterns)
        data["epoch"].append(i)
        data["train_loss"].append(mse_thresh(output_patterns, pred))
        logger.info("Epoch %d/%d. Train loss: %.4f", i, num_epochs,
                    data["train_loss"][-1])

        if data["train_loss"][-1] == 0:
            perfect_epochs += 1
        else:
            perfect_epochs = 0

        if perfect_epochs == 3:
            logger.info("Ending training after %d perfect epochs.",
                        perfect_epochs)
            break

    return pd.DataFrame(data)


def output(network: lb.Net, pattern: Iterable[float]) -> List[float]:
    """Calculates a prediction for a single input pattern.

    Args:
      network: The trained network.
      pattern: The input pattern.

    Returns:
      np.ndarray: The output of the network after clamping the input
      pattern to the input layer and settling. The max value is set to one,
      everything else is set to zero.

    """
    network.clamp_layer("input", pattern)
    for _ in range(50):
        network.cycle()
    network.unclamp_layer("input")
    out = network.observe("output", "unit_act")["act"].values
    return list(out)


def predict(network: lb.Net, input_patterns: np.ndarray) -> np.ndarray:
    """Calculates predictions for an array of input patterns.

    Args:
      network: The trained network.
      input_patterns: An array of shape (n_samples, n_features)
        containing the input patterns for which to calculate predictions.

    Returns:
      np.ndarray: An array of shape (n_samples, n_features) containing the
        predictions for the input patterns.

    """
    outputs = []
    for item in input_patterns:
        outputs.append(output(network, item))
    return np.array(outputs)


if __name__ == "__main__":
    PROJ_NAME = "err_hidden"

    logging.basicConfig(
        level=logging.DEBUG,
        format="%(asctime)s %(levelname)s %(message)s",
        handlers=(logging.FileHandler(
            "{0}_log.txt".format(PROJ_NAME), mode="w"),
                  logging.StreamHandler(sys.stdout)))

    logging.info("Begin training %s", PROJ_NAME)

    X, Y = load_data()
    net = build_network()
    metrics = train(net, X, Y)

    # Save metrics and network for future analysis
    metrics.to_csv("{0}_metrics.csv".format(PROJ_NAME), index=False)
    net.save("{0}_network.pkl".format(PROJ_NAME))
\end{minted}

  \subsection{Classifying the IRIS dataset}
\begin{minted}{python}
import logging
import sys
from typing import Any
from typing import Dict
from typing import List
from typing import Iterable
from typing import Tuple
from typing import Union

import numpy as np  # type: ignore
import pandas as pd  # type: ignore
import sklearn.datasets  # type: ignore
import sklearn.metrics  # type: ignore
import sklearn.model_selection  # type: ignore
import sklearn.preprocessing  # type: ignore

import leabra7 as lb  # type: ignore

LoggerType = Union[None, logging.Logger, logging.LoggerAdapter]


def load_data(num_feature_units: int = 10,
              logger: LoggerType = None) -> Tuple[np.ndarray, np.ndarray]:
    """Loads and preprocesses the data.

    Returns:
       An (X, Y) tuple containing the features and labels, respectively.

    """
    if logger is None:
        logger = logging.getLogger()
    logger.info("Loading data")
    data = sklearn.datasets.load_iris()

    # One-hot encode the labels
    label_binarizer = sklearn.preprocessing.LabelBinarizer()
    Y = label_binarizer.fit_transform(data.target)

    # Quantile transform, bin, and one-hot encode the features
    quant = sklearn.preprocessing.QuantileTransformer()
    X = quant.fit_transform(data.data)
    X = np.digitize(X, bins=np.linspace(0.0, 1.0, num=num_feature_units))
    one_hot = sklearn.preprocessing.OneHotEncoder(sparse=False)
    X = one_hot.fit_transform(X)

    # Randomly shuffle the data
    return sklearn.utils.shuffle(X, Y)


def build_network(input_size: int,
                  output_size: int,
                  hidden_size: int = 23,
                  logger: LoggerType = None) -> lb.Net:
    """Builds the classifier network.

    Args:
      input_size: The size of the input layer.
      output_size:  The size of the output layer.
      hidden_size:  The size of the hidden layer.
      logger:  The logger to use.

    Returns:
      A Leabra7 network for classification.

    """
    if logger is None:
        logger = logging.getLogger()
    logger.info("Building network")
    net = lb.Net()

    # Layers
    layer_spec = lb.LayerSpec(gi=1.5, ff=1, fb=1,
        unit_spec=lb.UnitSpec(spike_gain=0, vm_gain=0, adapt_dt=0))
    net.new_layer("input", size=input_size, spec=layer_spec)
    net.new_layer("hidden", size=hidden_size, spec=layer_spec)
    net.new_layer("output", size=output_size, spec=layer_spec)
    logger.debug("Input layer size: %d", input_size)
    logger.debug("Hidden layer size: %d", hidden_size)
    logger.debug("Output layer size: %d", output_size)

    # Projections
    lrate = 0.02
    up_spec = lb.ProjnSpec(
        lrate=lrate,
        dist=lb.Uniform(0.25, 0.75),
        cos_diff_thr_l_mix=False,
        cos_diff_lrate=False)
    down_spec = lb.ProjnSpec(
        lrate=lrate,
        dist=lb.Uniform(0.25, 0.5),
        wt_scale_rel=0.3,
        cos_diff_thr_l_mix=False,
        cos_diff_lrate=False)
    net.new_projn("input_to_hidden", pre="input", post="hidden", spec=up_spec)
    net.new_projn(
        "hidden_to_output", pre="hidden", post="output", spec=up_spec)
    net.new_projn(
        "output_to_hidden", pre="output", post="hidden", spec=down_spec)

    return net


def trial(network: lb.Net, input_pattern: Iterable[float],
          output_pattern: Iterable[float]) -> None:
    """Runs a trial.

    Args:
      input_pattern: The pattern to clamp to the network's input layer.
      output_pattern: The pattern to clamp to the network's output layer.

    """
    network.clamp_layer("input", input_pattern)
    network.minus_phase_cycle(num_cycles=50)
    network.clamp_layer("output", output_pattern)
    network.plus_phase_cycle(num_cycles=25)
    network.unclamp_layer("input")
    network.unclamp_layer("output")
    network.learn()


def epoch(network: lb.Net, input_patterns: np.ndarray,
          output_patterns: np.ndarray) -> None:
    """Runs an epoch (one pass through the whole dataset).

    Args:
      input_patterns: A numpy array with shape (n_samples, n_features).
      output_patterns: A numpy array with shape (n_samples, n_features).

    """
    for x, y in zip(input_patterns, output_patterns):
        trial(network, x, y)
    network.end_epoch()


def train(network: lb.Net,
          input_patterns: np.ndarray,
          output_patterns: np.ndarray,
          num_epochs: int = 500,
          logger: LoggerType = None) -> pd.DataFrame:
    """Trains the network.

    Args:
      input_patterns: A numpy array with shape (n_samples, n_features).
      output_patterns: A numpy array with shape (n_samples, n_features).
      num_patterns: The number of epochs to run. Defaults to 500.
      logger: The logger to use. If None, will use the module's default logger.

    Returns:
      pd.DataFrame:  A dataframe of metrics from the training run.

    """
    if logger is None:
        logger = logging.getLogger()
    logger.info("Begin training")

    X_train, X_test, Y_train, Y_test = sklearn.model_selection.train_test_split(
        input_patterns, output_patterns, test_size=0.2)

    logger.debug("Training set size: %d", X_train.shape[0])
    logger.debug("Test set size: %d", X_test.shape[0])

    data: Dict[str, List[float]] = {
        "epoch": [],
        "train_loss": [],
        "train_accuracy": [],
        "test_loss": [],
        "test_accuracy": []
    }

    for i in range(num_epochs):
        epoch(network, X_train, Y_train)

        # Predicting is slow, so we only calculate metrics every 5 epochs
        if (i + 1) % 5 == 0:
            pred_train = predict(network, X_train)
            data["epoch"].append(i)
            data["train_loss"].append(
                sklearn.metrics.mean_squared_error(Y_train, pred_train))
            data["train_accuracy"].append(
                sklearn.metrics.accuracy_score(
                    Y_train, pred_train, normalize=True))

            pred_test = predict(network, X_test)
            data["test_loss"].append(
                sklearn.metrics.mean_squared_error(Y_test, pred_test))
            data["test_accuracy"].append(
                sklearn.metrics.accuracy_score(
                    Y_test, pred_test, normalize=True))
            logger.info(
                "Epoch %d/%d. Train accuracy: %.2f%%. Test accuracy: %.2f%%",
                i, num_epochs, data["train_accuracy"][-1] * 100,
                data["test_accuracy"][-1] * 100)

    logger.info("End training")
    return pd.DataFrame(data)


def output(network: lb.Net, pattern: Iterable[float]) -> List[float]:
    """Calculates a prediction for a single input pattern.

    Args:
      network: The trained network.
      pattern: The input pattern.

    Returns:
      np.ndarray: The output of the network after clamping the input
      pattern to the input layer and settling. The max value is set to one,
      everything else is set to zero.

    """
    network.clamp_layer("input", pattern)
    for _ in range(50):
        network.cycle()
    network.unclamp_layer("input")
    out = network.observe("output", "unit_act")["act"].values
    max_idx = np.argmax(out)
    out[:] = 0
    out[max_idx] = 1
    return list(out)


def predict(network: lb.Net, input_patterns: np.ndarray) -> np.ndarray:
    """Calculates predictions for an array of input patterns.

    Args:
      network: The trained network.
      input_patterns: An array of shape (n_samples, n_features)
        containing the input patterns for which to calculate predictions.

    Returns:
      np.ndarray: An array of shape (n_samples, n_features) containing the
        predictions for the input patterns.

    """
    outputs = []
    for item in input_patterns:
        outputs.append(output(network, item))
    return np.array(outputs)


if __name__ == "__main__":
    PROJ_NAME = "iris"
    np.seterr("warn")

    logging.basicConfig(
        level=logging.DEBUG,
        format="%(asctime)s %(levelname)s %(message)s",
        handlers=(logging.FileHandler(
            "{0}_log.txt".format(PROJ_NAME), mode="w"),
                  logging.StreamHandler(sys.stdout)))

    logging.info("Begin training %s", PROJ_NAME)

    X, Y = load_data()
    net = build_network(input_size=X.shape[1], output_size=Y.shape[1])
    metrics = train(net, X, Y)

    # Save metrics and network for future analysis
    metrics.to_csv("{0}_metrics.csv".format(PROJ_NAME), index=False)
    net.save("{0}_network.pkl".format(PROJ_NAME))
\end{minted}

\end{document}